\documentclass{article}

\PassOptionsToPackage{numbers,compress}{natbib}

     \usepackage[preprint]{neurips_2020}

\usepackage[utf8]{inputenc} % allow utf-8 input
\usepackage[T1]{fontenc}    % use 8-bit T1 fonts
\usepackage{hyperref}       % hyperlinks
\usepackage{url}            % simple URL typesetting
\usepackage{booktabs}       % professional-quality tables
\usepackage{amsfonts}       % blackboard math symbols
\usepackage{nicefrac}       % compact symbols for 1/2, etc.
\usepackage{microtype}      % microtypography

\usepackage{amsmath}
\usepackage{graphicx}
\usepackage{algorithm}
\usepackage{algorithmic}
\usepackage{subcaption}
\usepackage[super]{nth}
\usepackage{siunitx}
\title{Unsupervised Learning of Solutions to Differential Equations with Generative Adversarial Networks}

\author{
    Dylan Randle \\
    IACS \\
    Harvard University \\
    Cambridge, MA 02138 \\
    \And
    Pavlos Protopapas \\
    IACS \\
    Harvard University \\
    Cambridge, MA 02138 \\
    \And
    David Sondak\\
    IACS \\
    Harvard University \\
    Cambridge, MA 02138 \\
}

\begin{document}

\maketitle

% =======================
%     ABSTRACT
% =======================
\begin{abstract}

Solutions to differential equations are of significant scientific and engineering relevance. Recently, there has been a growing interest in solving differential equations with neural networks. This work develops a novel method for solving differential equations with unsupervised neural networks that applies Generative Adversarial Networks (GANs) to \emph{learn the loss function} for optimizing the neural network. We present empirical results showing that our method, which we call Differential Equation GAN (DEQGAN), can obtain multiple orders of magnitude lower mean squared errors than an alternative unsupervised neural network method based on (squared) $L_2$, $L_1$, and Huber loss functions. Moreover, we show that DEQGAN achieves solution accuracy that is competitive with traditional numerical methods. Finally, we analyze the stability of our approach and find it to be sensitive to the selection of hyperparameters, which we provide in the \hyperref[appendix]{Appendix}.\footnote{Code available at \url{https://github.com/dylanrandle/denn}. Please address any electronic correspondence to \texttt{dylanrandle@alumni.harvard.edu}.}

\end{abstract}

% =======================
%     INTRODUCTION
% =======================
\section{Introduction}
\label{intro}

In fields such as physics, chemistry, biology, engineering, and economics, differential equations are applied to the modeling of important and complex phenomena. While traditional methods for solving differential equations perform well, and the theory for their stability and convergence are well established, the recent success of deep learning techniques \cite{imagenet, seq2seq, attention, transformer, RL_atari, RL_dist, RL_robotic_manipulation, RL_Go} has inspired researchers to apply neural networks to solving differential equations \cite{physics_informed_nns, rnns_odesystem, denns_cosmos, mattheakis2019physical, conv_lstm_pdes, mattheakis2020hamiltonian, nn_highdim_pdes, highdim_nn_pde_forward_backward, dgm_highdim_pde_nn}. 

Applying neural networks to solving differential equations can provide a range of benefits over traditional methods. By removing a reliance on finely crafted grids which suffer from the ``curse of dimensionality", neural networks can be more effective than traditional solvers in high-dimensional settings \cite{nn_highdim_pdes, highdim_nn_pde_forward_backward, dgm_highdim_pde_nn}. Furthermore, recent work has shown that neural network solutions can be more accurate in obeying certain physical constraints, such as conservation of energy \cite{mattheakis2020hamiltonian, physics_informed_nns}. Neural networks can also provide provide a more accurate interpolation scheme \cite{Lagaris_1998}. Finally, forward passes of neural networks are embarrassingly data-parallel, even in difficult-to-parallelize temporal dimensions, and can readily leverage parallel computing architectures.

Interest in research on solving differential equations with unsupervised neural networks is growing. However, due to a lack (to the best of our knowledge) of theoretical justification for a particular choice of loss function, we propose ``learning" the loss function with a Generative Adversarial Network (GAN) \cite{goodfellow2014generative}. For data following a known noise model, there is clear theoretical justification, based on the maximum likelihood principle, for fitting models with particular loss functions. For example, in the case of a Gaussian noise model
\begin{equation}
    y = x + \epsilon, \quad \epsilon \sim \mathcal{N}(0, \sigma^2),
\end{equation}
the maximum likelihood estimate of the model parameters minimizes the squared error ($L_2$ norm) loss function. In the case of deterministic differential equations, however, there is no noise model and we lack formal justification for a particular choice of loss function among multiple options. 

To circumvent this problem, we propose GANs for solving differential equations in a fully unsupervised manner. The discriminator model of a GAN can be thought of as learning the loss function used for optimizing the generator. Moreover, GANs have been shown to excel in scenarios where classic loss functions, such as the mean squared error, struggle due to their inability to capture complex spatio-temporal dependencies \cite{GAN_VAE, superresolution, styleGAN}. 

Our main contribution is a novel method, which we call Differential Equation GAN (DEQGAN), for formulating the task of solving differential equations in a \emph{fully unsupervised} manner as a GAN training problem. DEQGAN works by separating the differential equation into left-hand side ($LHS$) and right-hand side ($RHS$), then training the generator to produce a $LHS$ that is indistinguishable to the discriminator from the $RHS$. Experimental results show that our method produces solutions which obtain multiple orders of magnitude lower mean squared errors (computed from known analytic or numerical solutions) than a comparable unsupervised neural network method with (squared) $L_2$, $L_1$, and Huber loss functions. Moreover, DEQGAN achieves solution accuracy that is competitive with traditional fourth-order Runge-Kutta and second-order finite difference methods.

% =======================
%     RELATED WORK
% =======================
\section{Related Work}

\citet{dissanayake1994neural} were one of the first to develop a method for solving differential equations with neural networks. They showed that a neural network-based method could solve differential equations when transformed to unconstrained optimization problems. \citet{Lagaris_1998} extended this work by introducing analytical adjustments to the neural network output that exactly satisfied initial and boundary conditions. They showed that their method achieved lower interpolation error than the finite element method, while maintaining equal error on a fixed mesh. The authors expanded this work to consider arbitrarily-shaped domains in higher dimensions \cite{lagaris_arbitrary_boundary}, and applied neural networks to quantum mechanics \cite{Lagaris_quantum}.

Recent work has solved high-dimensional partial differential equations (PDEs) with neural networks in place of basis functions \cite{dgm_highdim_pde_nn} and by reformulating PDEs as backward stochastic differential equations \cite{nn_highdim_pdes}. To reduce the need to re-learn known physics, \citet{mattheakis2019physical} embedded physical symmetries into the structure of neural networks, and \citet{physics_informed_nns} regularized neural networks according to physical models described by nonlinear PDEs, leading to improved solution accuracy and training convergence over physics-agnostic counterparts. Leveraging recent advances in deep learning, \citet{rnns_odesystem} developed a supervised recurrent neural network method that uses measurement data to solve ordinary differential equations with unknown functional forms, and \citet{conv_lstm_pdes} presented a fully convolutional LSTM network that augments traditional finite difference/finite volume methods used to solve PDEs. \citet{nn4diffeq_survey} presented a survey of neural network and radial basis function methods for solving differential equations.

In parallel, \citet{goodfellow2014generative} introduced the idea of learning generative models with neural networks and an adversarial training algorithm, called Generative Adversarial Networks (GANs). To solve issues of GAN training instability, \citet{arjovsky2017wasserstein} introduced a formulation of GANs based on the Wasserstein distance, and \citet{gulrajani2017improved} added a gradient penalty to approximately enforce a Lipschitz constraint on the discriminator. \citet{gan_spectral_norm} introduced an alternative method for enforcing the Lipschitz constraint with a spectral normalization technique that outperforms the former method on some problems.

Further work has applied GANs to differential equations with solution data used for supervision. \citet{physicsinformedGAN} apply GANs to stochastic differential equations by using ``snapshots" of ground-truth data for semi-supervised training. A project by students at Stanford \cite{turbulence_enrichment} employed GANs to perform ``turbulence enrichment" of solution data in a manner akin to that of super-resolution for images proposed by \citet{superresolution}. Our work distinguishes itself from other GAN-based approaches for solving differential equations by being \emph{fully unsupervised}, and removing the dependence on using supervised training data (i.e. solutions of the equation).

% =======================
%     BACKGROUND
% =======================
\section{Background}
\label{sec:background}

\subsection{Unsupervised Neural Networks for Differential Equations}

Early work by \citet{dissanayake1994neural} proposed solving differential equations in an unsupervised manner with neural networks. Their paper considers general differential equations of the form

\begin{equation}  \label{eq:lagaris_diffeq}
F(t, \Psi(t), \Delta \Psi(t), \Delta^2 \Psi(t), \ldots) = 0
\end{equation}
where $\Psi(t)$ is the desired solution, $\Delta$ and $\Delta^2$ represent the first and second derivatives, and the system is subject to certain boundary or initial conditions. The learning problem is then formulated as minimizing the sum of squared errors of the above equation
\begin{equation}  \label{eq:lagaris_objective}
    \min_{\theta}{\sum_{t \in \mathcal{D}}{F(t, \Psi_{\theta}(t), \Delta \Psi_{\theta}(t), \Delta^2 \Psi_{\theta}(t), \ldots)^2}}
\end{equation}
where $\Psi_{\theta}$ is a neural network parameterized by $\theta$, $\mathcal{D}$ is the domain of the problem, and we compute derivatives with automatic differentiation. This allows us to use backpropagation \cite{backprop} to train the parameters of the neural network to satisfy the differential equation. Note that this formalism can be trivially extended to handle spatial domains and multidimensional problems, which we do in our experiments and describe in the \hyperref[appendix]{Appendix}.

\subsection{Generative Adversarial Networks}

Generative Adversarial Networks (GANs) \cite{goodfellow2014generative} are a type generative model that use two neural networks to induce a generative distribution $p(x)$ of the data by formulating the inference problem as a two-player, zero-sum game. 

The generative model first samples a latent variable $z \sim \mathcal{N}(0,1)$, which is used as input into the generator $G$ (e.g. a neural network). A discriminator $D$ is trained to classify whether its input was sampled from the generator (i.e. ``fake") or from a reference data set (i.e. ``real").

Informally, the process of training GANs proceeds by optimizing a minimax objective over the generator and discriminator such that the generator attempts to trick the discriminator to classify ``fake" samples as ``real". Formally, one optimizes

\begin{equation} \label{eq:gan_goodfellow}
    \min_{G} \max_{D} V(D,G) = \mathbb{E}_{x \sim p_{\text{data}}(x)}[\log{D(x)}]
    + \mathbb{E}_{z \sim p_{z}(z)}[1 - \log{D(G(z))}]
\end{equation}
where $x \sim p_{\text{data}}(x)$ denotes samples from the empirical data distribution, and $p_z \sim \mathcal{N}(0,1)$ samples in latent space. In practice, the optimization alternates between gradient ascent and descent steps for $D$ and $G$ respectively. 

\subsubsection{Two Time-Scale Update Rule}

\citet{ttur_gan} proposed the two time-scale update rule (TTUR) for training GANs, a method in which the discriminator and generator are trained with separate learning rates. They showed that their method led to improved performance and proved that, in some cases, TTUR ensures convergence to a stable local Nash equilibrium. One intuition for TTUR comes from the potentially different loss surface curvatures of the discriminator and generator. Allowing learning rates to be tuned to a particular loss surface can enable more efficient gradient-based optimization. We make use of TTUR throughout this paper as an instrumental lever when tuning GANs to reach desired performance.

\subsubsection{Spectral Normalization}

Proposed by \citet{gan_spectral_norm}, Spectrally Normalized GAN (SN-GAN) is a method for controlling exploding discriminator gradients when optimizing Equation \ref{eq:gan_goodfellow} that leverages a novel weight normalization technique. 

The key idea is to control the Lipschitz constant of the discriminator by constraining the spectral norm of each layer in the discriminator. Specifically, the authors propose dividing the weight matrices  $W_i$ of each layer $i$ by their spectral norm $\sigma(W_i)$

\begin{equation}
    W_{SN,i} = \frac{W_i}{\sigma(W_i)},
\end{equation}
where

\begin{equation}
    \sigma(W_i) = \max_{\| h_i\|_2 \leq 1} \| W_i h_i \|_2 
\end{equation}
and $h_i$ denotes the input to layer $i$. The authors prove that this normalization technique bounds the Lipschitz constant of the discriminator above by $1$, thus strictly enforcing the $1$-Lipshcitz constraint on the discriminator. In our experiments, adopting the SN-GAN formulation leads to even better performance than WGAN-GP \cite{arjovsky2017wasserstein, gulrajani2017improved}.

\subsection{Guaranteeing Initial \& Boundary Conditions} \label{physical_constraints}
 \citet{Lagaris_1998} showed that it is possible to exactly satisfy initial and boundary conditions by adjusting the output of the neural network. For example, consider adjusting the neural network solution $\Psi(t)$ to satisfy the initial condition $\Psi(0) = x_0$. We can apply the transformation \begin{equation}
     \Psi(t)' = x_0 + t \Psi(t)
 \end{equation} which exactly satisfies the condition. \citet{mattheakis2019physical} proposed an augmented transformation
\begin{equation} \label{eq:adjustment}
    \Psi(t)' = \Phi\left(\Psi(t)\right) = x_0 + \left(1 - e^{-(t-t_0)}\right) \Psi(t)
\end{equation}
that further improved training convergence. Intuitively, Equation \ref{eq:adjustment} adjusts the output of the neural network $\Psi(t)$ to be exactly $x_0$ when $t=t_0$, and decays this constraint exponentially in $t$.

\subsection{Residual Connections}
\citet{residual_connections} showed  that adding residual connections improved training of deep neural networks. We employ residual connections to our deep networks as they allow gradients to more easily flow through the models and thereby reduce numerical instability. Residual connections augment a typical activation with the identity operation

\begin{equation}
    y = \mathcal{F}(x, W_i) + x
\end{equation}
where $\mathcal{F}$ is the activation function, $x$ is the input to the unit, $W_i$ are the weights, and $y$ is the output of the unit. This acts as a ``skip connection", allowing inputs and gradients to forego the nonlinear component.

% =======================
%     METHOD
% =======================

\section{Differential Equation GAN}

Here we present our method, Differential Equation GAN (DEQGAN), which trains a GAN to solve differential equations in a \emph{fully unsupervised} manner. To do this, we rearrange the differential equation such that the left-hand side ($LHS$) contains all of the terms which depend on the generator (e.g. $\hat{\Psi}$, $\Delta \hat{\Psi}$, $\Delta^2 \hat{\Psi}$, etc.), and the right-hand side ($RHS$) contains only constants (e.g. zero). 

From here we sample points from the domain $t \sim \mathcal{D}$ and use them as input to a generator $G(t)$, which produces candidate solutions $\hat{\Psi}$. We adjust $\hat{\Psi}$ for initial or boundary conditions according to Equation~\ref{eq:adjustment}. Then we construct the $LHS$ from the differential equation $F$ using automatic differentiation
\begin{equation}
    LHS = F\left(t, \hat{\Psi}(t), \Delta \hat{\Psi}(t), \Delta^2 \hat{\Psi}(t)\right)
\end{equation}
and set $RHS$ to its appropriate value (in our examples, $RHS=0$).

\begin{figure}
  \centering
  \includegraphics[width=0.9\textwidth]{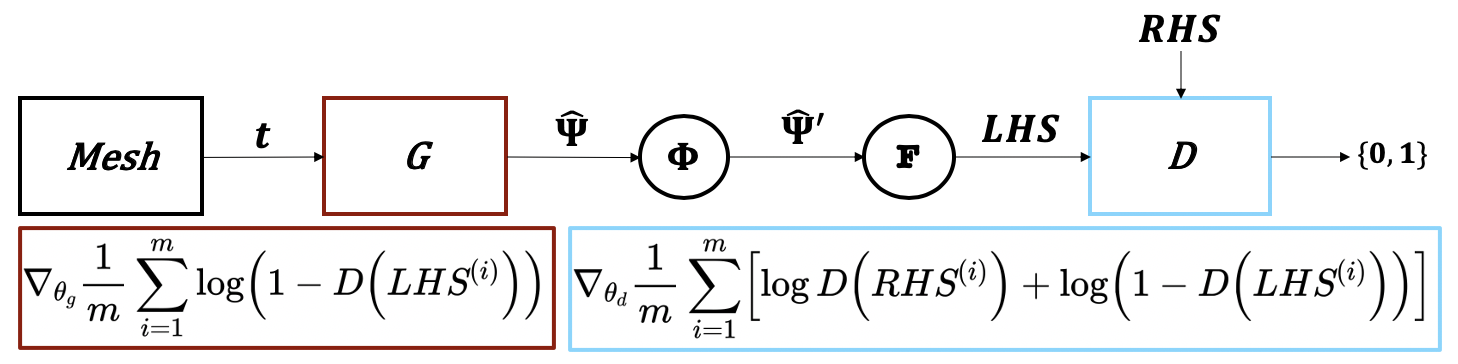}
  \caption[DEQGAN Diagram]{Schematic representation of DEQGAN. We perturb points $t$ from the mesh and input them to a generator $G$, which produces candidate solutions $\hat{\Psi}$. Then we analytically adjust these solutions according to $\Phi$ and apply automatic differentiation to construct $LHS$ from the differential equation $F$. $RHS$ and $LHS$ are passed to a discriminator $D$, which is trained to classify them as ``real" and ``fake" respectively.}
  \label{fig:gan_diagram}
\end{figure}

From here, training proceeds in a manner similar to traditional GANs. We update the weights of the generator $G$ and discriminator $D$ according to the gradients

\begin{equation}
\label{eq:generator_grad}
    \eta_G = \nabla_{\theta_{g}} \frac{1}{m} \sum_{i=1}^{m} \log{ \left(1 - D \left( LHS^{(i)} \right) \right)},
\end{equation}
\begin{equation} \label{eq:discriminator_grad}
    \eta_{D} = \nabla_{\theta_{d}} \frac{1}{m} \sum_{i=1}^{m} \left[ \log D \left( RHS^{(i)} \right) +  
    \log \left( 1 - D \left( LHS^{(i)} \right) \right) \right]
\end{equation}
where $LHS^{(i)}$ is the output of $G\left(t^{(i)}\right)$ after adjusting for initial or boundary conditions and constructing the $LHS$ from $F$. Note that we perform stochastic gradient \emph{descent} for $G$ (gradient steps $\propto -\eta_{G}$), and stochastic gradient \emph{ascent} for $D$ (gradient steps $\propto \eta_{D}$). We provide a schematic representation of DEQGAN in Figure \ref{fig:gan_diagram} and detail the training steps in Algorithm \ref{alg:gan_algo}.

\begin{algorithm}
    \caption{DEQGAN}
    \label{alg:gan_algo}
    \begin{algorithmic}
    \STATE {\bfseries Input:} Differential equation $F$, generator $G(\cdot;\theta_g)$, discriminator $D(\cdot;\theta_d)$, mesh $t$ of $m$ points with spacing $\Delta t$, perturbation precision $\tau$, analytic adjustment function $\Phi$, total steps $N$, learning rates $\alpha_G, \alpha_D$, Adam optimizer \cite{adamoptimizer2014} parameters $\beta_{G1}, \beta_{G2}, \beta_{D1}, \beta_{D2}$ \\
    \FOR{$i=1$ {\bfseries to} $N$}
        \FOR{$j=1$ {\bfseries to} $m$}
        \STATE Perturb $j$-th point in mesh $t_{s}^{(j)} = t^{(j)} + \epsilon, \epsilon \sim \mathcal{N}(0,\frac{\Delta t}{\tau})$
        \STATE Forward pass $\hat{\Psi} = G(t_{s}^{(j)})$
        \STATE Analytic adjustment $\hat{\Psi}' = \Phi(\hat{\Psi})$ (Equation \ref{eq:adjustment})
        \STATE Compute $LHS^{(j)} = F(t, \hat{\Psi}', \nabla{\hat{\Psi}'}, \nabla^{2}{\hat{\Psi}'})$, set $RHS^{(j)} = 0$ 
        \ENDFOR
    \STATE Compute gradients $\eta_G, \eta_D$ (Equation \ref{eq:generator_grad} \& \ref{eq:discriminator_grad})
    \STATE Update generator $\theta_g \leftarrow \texttt{Adam}(\theta_g, -\eta_G, \alpha_G, \beta_{G1}, \beta_{G2})$
    \STATE Update discriminator $\theta_d \leftarrow \texttt{Adam}(\theta_d, \eta_D, \alpha_D, \beta_{D1}, \beta_{D2})$
    \ENDFOR
    \STATE {\bfseries Output:} $G$
    \end{algorithmic}
\end{algorithm}

Informally, our algorithm trains a GAN by setting the ``fake" component to be the $LHS$ (in our formulation, the residuals of the equation), and the ``real" component to be the $RHS$ of the equation. This results in a GAN that learns to produce solutions that make $LHS$ indistinguishable from $RHS$, thereby approximately solving the differential equation.

An important note here is that training can be unstable if $LHS$ and $RHS$ are not chosen properly. Specifically, we find that training fails if $RHS$ is a function of the generator. For example, consider the equation $\ddot{x}+x=0$. If we set $LHS=\ddot{x}$ and $RHS=-x$, then RHS is a function of the generator and will be constantly changing as the generator is updated throughout training, and DEQGAN will become exceedingly unstable. We can fix this, however, by simply setting $LHS=\ddot{x}+x$ and $RHS=0$. 

Our intuition for this is that if $RHS$ depends on the outputs of the generator, the ``real" data distribution $p_{\text{data}}(x)$ (from Equation \ref{eq:gan_goodfellow}) changes as the generator weights are updated throughout training. If the distribution $p_{\text{data}}(x)$ is constantly changing, the discriminator will not have a reliable signal for learning to classify ``real" from ``fake", which violates a core assumption of traditional GANs. By setting $RHS=0$, we resolve the problem by effectively setting the ``real" distribution to be the fixed Dirac delta function $p_{\text{data}}(x) = \delta(0)$. For the examples in this paper, we move all terms of the differential equation to $LHS$ and set $RHS=0$.

% =======================
%     EXPERIMENTS
% =======================
\begin{table}
  \caption{Summary of Experiments}
  \label{table:summary_of_experiments}
  \centering
  \begin{tabular}{lccccc}
    \toprule
    Key & Equation & Class & Order & Linear & System \\ 
    \midrule
    EXP & $\dot{x}(t) + x(t) = 0$ & ODE & \nth{1} & Yes & No \\
    SHO & $\ddot{x}(t) + x(t) = 0$ &  ODE & \nth{2} & Yes & No \\
    NLO & $\begin{aligned}
    \ddot{x}(t)+2 \beta \dot{x}(t)+\omega^{2} x(t)
    +\phi x(t)^{2}+\epsilon x(t)^{3} = 0
    \end{aligned}$ & ODE & \nth{2} & No & No \\
    NAS & $\begin{cases}
    \dot{x}(t) = -ty \\
    \dot{y}(t) = tx
    \end{cases}$ & ODE & \nth{1} & Yes & Yes  \\
    SIR & $\begin{cases}
    \dot{S}(t) &= - \beta I(t)S(t)/N \\
    \dot{I}(t) &= \beta I(t)S(t)/N - \gamma I(t) \\
    \dot{R}(t) &= \gamma I(t)
\end{cases}$ & ODE & \nth{1} & No & Yes \\
    POS & $\begin{aligned} u_{xx} + u_{yy} = 2x(y-1)(y-2x+xy+2)e^{x-y}
    \end{aligned}$ & PDE & \nth{2} & Yes & No \\
    \bottomrule
  \end{tabular}
\end{table}

\begin{table}
  \caption{Experimental Results}
  \label{table:experimental_results}
  \centering
  \begin{tabular}{lccccccccc}
    \toprule
    & \multicolumn{5}{c}{Mean Squared Error} \\ 
    \cmidrule(lr){2-6}
    Key & $L_1$ & $L_2$ & Huber & DEQGAN & Traditional  \\ 
    \midrule
    EXP & \num{1e-3} & \num{3e-6} & \num{1e-6} & \num{3e-16} & \num{2e-14} (RK4) \\
    SHO & \num{2e-5} & \num{2e-9} & \num{8e-10} & \num{1e-12} & \num{1e-11} (RK4) \\
    NLO & \num{6e-2} & \num{1e-9} & \num{4e-10} & \num{2e-12} & \num{4e-11} (RK4) \\
    NAS & \num{6e-1} & \num{6e-5} & \num{2e-3} & \num{8e-9} & \num{2e-9} (RK4) \\
    SIR & \num{7e-4} & \num{6e-9} & \num{3e-9}  & \num{2e-10}  & \num{5e-13} (RK4) \\
    POS  & \num{4e-6} & \num{5e-11} & \num{2e-11} & \num{8e-13} & \num{3e-10} (FD) \\
    \bottomrule
  \end{tabular}
\end{table}

\section{Experiments}

We conducted experiments on several differential equations of increasing complexity, comparing DEQGAN to an alternative unsupervised neural network method using squared $L_2$ (i.e. mean squared error\footnote{We use the term $L_2$ to avoid conflating the \emph{loss function} being used, which is the mean squared error on the unsupervised problem of minimizing the differential equation residuals, with the final \emph{evaluation metric}, which is the mean squared error of the predicted solution computed against the known ground truth.}), $L_1$, and Huber \cite{huber_loss} loss functions. We also report results obtained by the traditional fourth-order Runge-Kutta (RK4) and second-order finite differences (FD) methods for initial and boundary value problems, respectively. A detailed description of each experiment including exact problem specifications, hyperparameters, and a comparison of loss functions used is provided in the \hyperref[appendix]{Appendix}.

We report the mean squared errors of the solutions produced by each method, computed from known solutions obtained either analytically or with high-quality numerical solvers \cite{scipy_cite}, but we do not use these solution data for training. We add residual connections between neighboring layers of all models, and apply spectral normalization to the discriminator in all experiments. Results are obtained with hyperparameters tuned for DEQGAN. In the \hyperref[appendix]{Appendix}, we tune each alternative method for comparison but do not observe a meaningful difference.

We note that deterministic differential equations do not exhibit aleatoric uncertainty and that all errors we observe are therefore epistemic. Given that neural networks are, theoretically, universal function approximators \cite{universalapproximator}, one may initially expect to obtain arbitrarily low error. The reason we do not observe this is, first, that our models are finite-width and may lack representational capacity and, second, that optimizing the objective in Equation \ref{eq:gan_goodfellow} with stochastic gradient descent is unlikely to reach globally optimal solutions and is more likely to converge to local optima.

\begin{figure}
    \centering
    \begin{subfigure}[b]{0.475\textwidth}
        \centering
        \includegraphics[width=\textwidth]{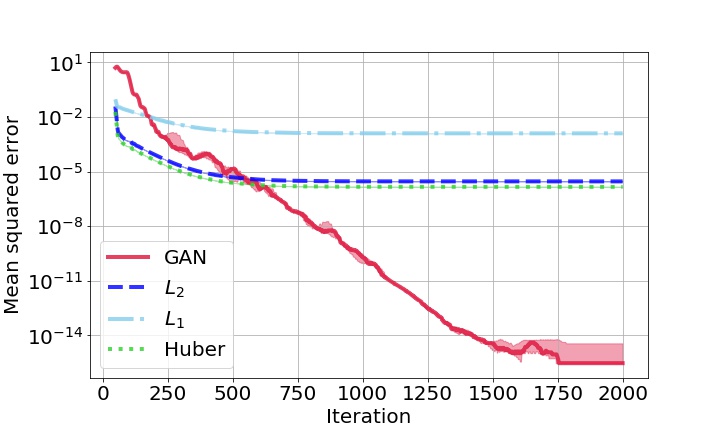}
        \caption{Exponential Decay (EXP)}
        \label{fig:exp_comparison}
    \end{subfigure}
    \hfill
    \begin{subfigure}[b]{0.475\textwidth}  
        \centering 
        \includegraphics[width=\textwidth]{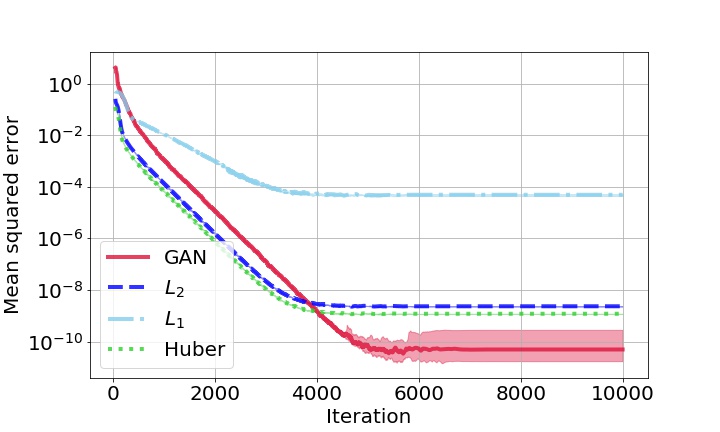}
        \caption{Simple Harmonic Oscillator (SHO)}    
        \label{fig:sho_gan_vs_l2}
    \end{subfigure}
    \vskip\baselineskip
    \begin{subfigure}[b]{0.475\textwidth}   
        \centering 
        \includegraphics[width=\textwidth]{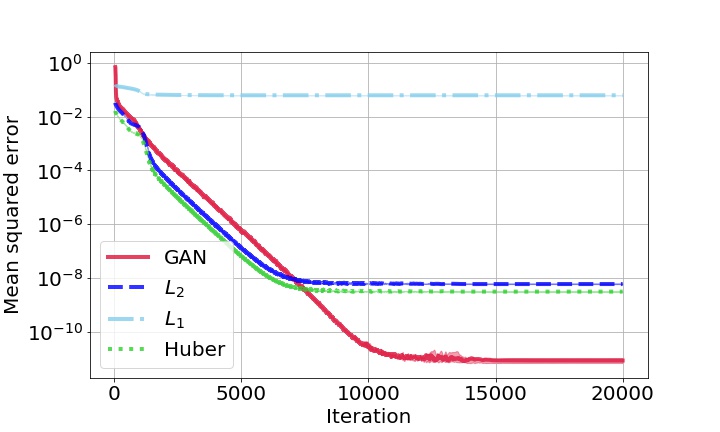}
        \caption{Nonlinear Oscillator (NLO)}
        \label{fig:nlo_gan_vs_l2}
    \end{subfigure}
    \quad
    \begin{subfigure}[b]{0.475\textwidth}   
        \centering 
        \includegraphics[width=\textwidth]{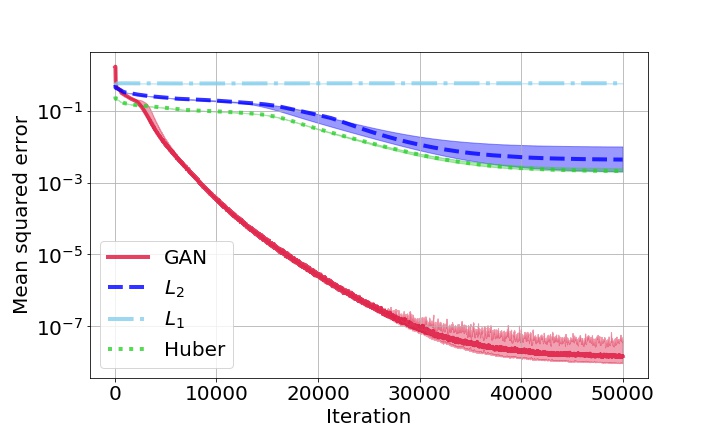}
        \caption{Non-Autonomous System (NAS)}
        \label{fig:coo_gan_vs_l2}
    \end{subfigure}
    \vskip\baselineskip
    \begin{subfigure}[b]{0.475\textwidth}   
        \centering 
        \includegraphics[width=\textwidth]{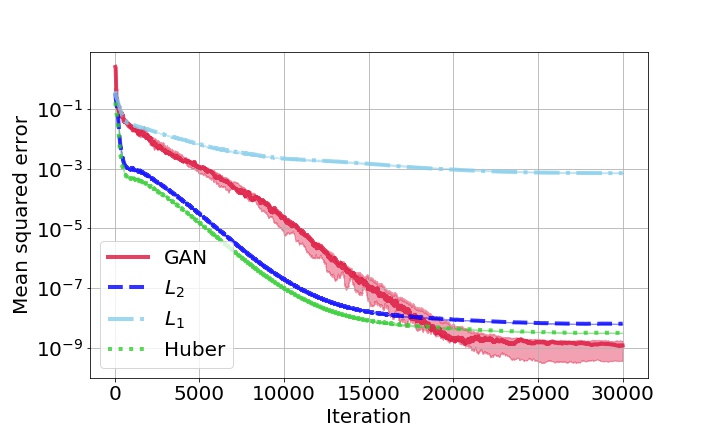}
        \caption{SIR Disease Model (SIR)}    
        \label{fig:sir_gan_vs_l2}
    \end{subfigure}
    \quad
    \begin{subfigure}[b]{0.475\textwidth}   
        \centering 
        \includegraphics[width=\textwidth]{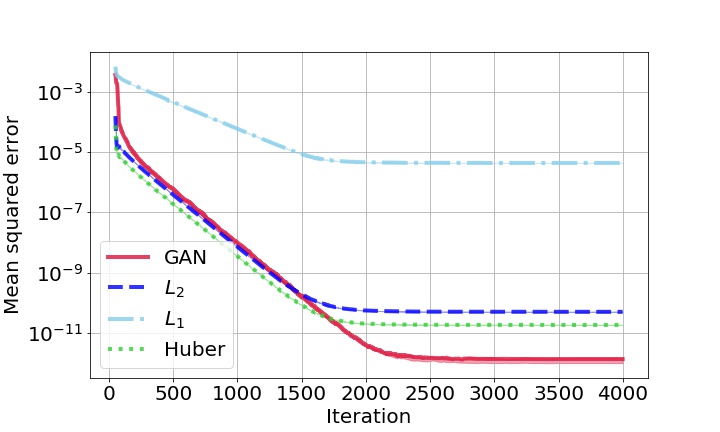}
        \caption{Poisson Equation (POS)}    
        \label{fig:pos_gan_vs_l2}
    \end{subfigure}
    \caption{Mean squared errors vs. iteration for DEQGAN, $L_2$,  $L_1$, and Huber loss for various equations. We perform five randomized trials and plot the median (bold) and $(25, 75)$ percentile range (shaded). We smooth the values using a simple moving average with window size $50$.}
    \label{fig:experiments_rand_reps}
\end{figure}

Table \ref{table:summary_of_experiments} summarizes the equations we study in our experiments, and Table \ref{table:experimental_results} reports the lowest mean squared errors obtained across five trials for each method. We see that DEQGAN obtains multiple orders of magnitude lower mean squared errors than the alternative unsupervised neural network method with $L_1$, $L_2$, and Huber loss functions across the differential equations studied. Moreover, DEQGAN achieves accuracy that is competitive with the traditional RK4 and FD numerical methods.

Figure \ref{fig:experiments_rand_reps} plots mean squared error against training iteration for DEQGAN and the alternative neural network method with $L_1$, $L_2$ and Huber loss functions. We observe that DEQGAN converges to lower mean squared errors than the alternative unsupervised neural network method, often by multiple orders of magnitude, across the equations studied. We note, however, that the mean squared error curves of DEQGAN are less smooth and exhibit greater variability than the normal unsupervised neural network method, an issue we study in the following section.

% =======================
%     DISCUSSION
% =======================
\section{Stability of DEQGAN Training}
\label{discussion}

A point that we have not addressed is the instability of the DEQGAN training algorithm. The instability of GANs is not a new problem and much work has been dedicated to improving the stability and convergence of GANs \cite{arjovsky2017wasserstein, gulrajani2017improved, BEGAN, mirza2014conditional, gan_spectral_norm}. In our experiments we find that the initial weights of the generator and discriminator can have a substantial impact on the final performance of DEQGAN. The solution that we adopt is to fix the initial model weights when tuning hyperparameters for DEQGAN and to keep the same weight initialization thereafter, which appears to significantly reduce the problem of instability.

\begin{figure}[h]
  \centering
  \includegraphics[width=0.9\textwidth]{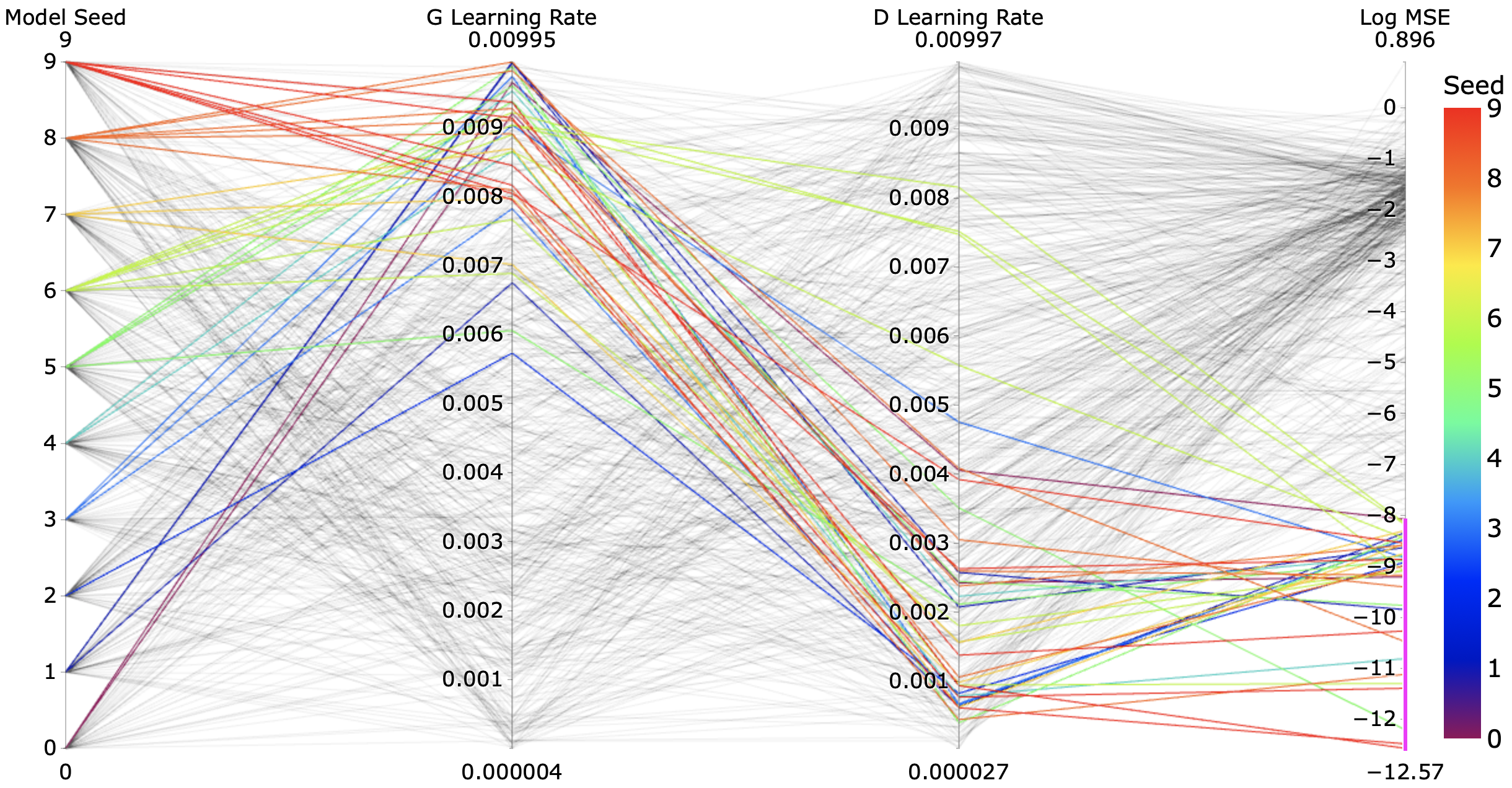}
  \caption[DEQGAN, Initialization, and Hyperparameters: Best MSEs]{Parallel plot showing the results of $500$ DEQGAN experiments on the exponential decay equation. The colors represent the different random seeds used to initialize model weights. We filter the results (non-selected lines appear gray) to highlight experiments achieving mean squared error $\leq 10^{-8}$. The mean squared error is plotted on a $log_{10}$ scale.}
  \label{fig:exp_seeds_mse}
\end{figure}

To illustrate the relationship between performance, hyperparameters, and initial model weights, Figure \ref{fig:exp_seeds_mse} plots the results of $500$ DEQGAN experiments for the exponential decay equation. For each experiment, we uniformly at random select model weight initialization random seeds as integers from the range $[0,9]$, as well as separate learning rates for the discriminator and generator in the range $[10^{-6}, 10^{-2}]$. We then record the final mean squared error on the validation set after running DEQGAN training for $500$ steps. Each line represents a combination of model weight initialization random seed, learning rate hyperparameters, and final (log) mean squared error of a single experiment. 

Notably, the results as a whole exhibit considerable variation in final mean squared error. However, when we filter on experiments achieving low mean squared errors ($\leq 10^{-8}$), we see that hyperparameter settings exist, for each of the model weight initialization seeds, that provide highly accurate DEQGAN solutions. We also observe a pattern in the hyperparameters which produce the low mean squared error experiments. We note that relatively high generator learning rates and low discriminator learning rates lead to the best DEQGAN performance across different model initialization seeds.

% =======================
%     CONCLUSION
% =======================
\section{Conclusion}
\label{conclusion}

We have presented a novel method which leverages GAN-based adversarial training to ``learn" the loss function for solving differential equations with unsupervised neural networks. We have shown empirically that our our method, which we call Differential Equation GAN (DEQGAN), can obtain multiple orders of magnitude lower mean squared errors than an alternative unsupervised neural network method with $L_2$, $L_1$, and Huber loss functions. Moreover, we show that DEQGAN achieves solution accuracy that is competitive with the traditional fourth-order Runge-Kutta and second-order finite difference methods. While our approach is sensitive to hyperparmaters, we have shown that it is possible to train a GAN in a \emph{fully unsupervised} manner to achieve highly accurate solutions to differential equations.

\newpage

% BROADER IMPACT
\section*{Broader Impact}
We hope that the broader impact of this work will be to advance the study of unsupervised neural network methods for solving differential equations. We do not believe that our work holds particularly poignant ethical or societal consequences. We note that our method does not provide theoretical guarantees of solution accuracy, and any critical implementations relying on this should exercise caution.

\begin{ack}
The authors would like to acknowledge helpful discussions with Marios Mattheakis, Cengiz Pehlevan, and Feiyu Chen.
\end{ack}

% REFERENCES
\bibliographystyle{apalike2}
\bibliography{references}

\newpage

% APPENDIX
\section*{Appendix}
\label{appendix}

\subsection*{Description of Experiments}
\label{appendix:experiments}

\begin{figure}[h]
    \centering
    \includegraphics[width=0.4\textwidth]{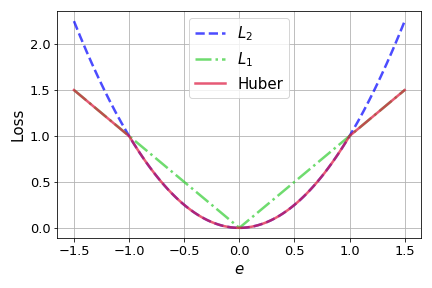}
    \caption[Comparison of Classic Loss Functions]{Comparison of $L_2$, $L_1$, and Huber loss functions. The Huber loss is equal to $L_2$ for $e \leq 1$ and to $L_1$ for $e > 1$.}
    \label{fig:huber_loss_l1_l2_simple}
\end{figure}

A plot of the various classical loss functions is provided in Figure \ref{fig:huber_loss_l1_l2_simple}.

% EXPONENTIAL
\paragraph{Exponential Decay (EXP)} \label{sec:exp_decay}
Consider a model for population decay $x(t)$ given by the exponential differential equation
\begin{equation}
    \dot{x}(t) + x(t) = 0, 
\end{equation} 
with $ x(0) = 1 $ and $t \in (0, 10)$. The ground truth solution $x(t) = e^{-t}$ can be obtained analytically. We reiterate, however, that our method is fully unsupervised and does not make use of this solution data during training. We simply use the ground truth solutions to report mean squared errors of predicted solutions.

\begin{figure}[h]
  \centering
  \includegraphics[width=\textwidth]{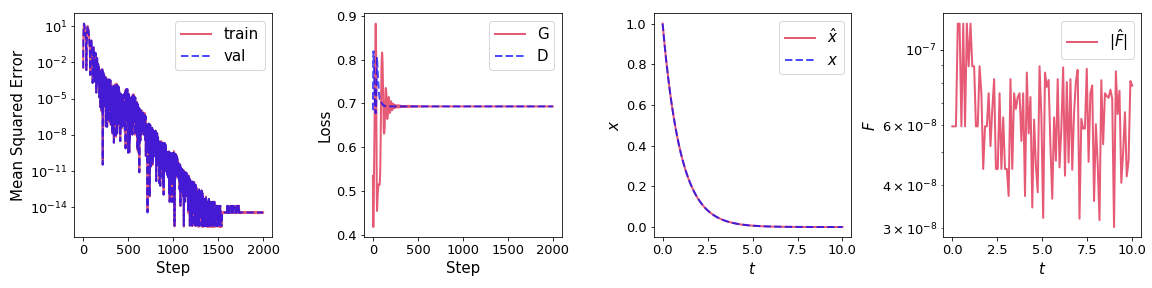}
  \caption[DEQGAN Training: Exponential Decay]{Visualization of DEQGAN training for the exponential decay problem. The left-most figure plots the mean squared error vs. iteration. To the right, we plot the value of the generator (G) and discriminator (D) losses at each iteration. Right of this we plot the prediction of the generator $\hat{x}$ and the true analytic solution $x$ as functions of time $t$. The right-most figure plots the absolute value of the residual of the predicted solution $\hat{F}$.}
  \label{fig:gan_exp}
\end{figure}

To set up the problem for DEQGAN, we define $LHS = \dot{x} + x$ and $RHS = 0$. Figure \ref{fig:gan_exp} presents the results from training DEQGAN on this equation. 

% SIMPLE HARMONIC OSCILLATOR
\paragraph{Simple Harmonic Oscillator (SHO)}
Consider the motion of an oscillating body $x(t)$, which can be modeled by the simple harmonic oscillator differential equation
\begin{equation}
    \ddot{x}(t) + x(t) = 0,
\end{equation}
with $x(0) = 0$,  $ \dot{x}(0) = 1$, and $t \in (0, 2\pi)$. This differential equation can be solved analytically and has an exact solution $x(t) = \sin{t}$. 

\begin{figure}[h]
  \centering
  \includegraphics[width=\textwidth]{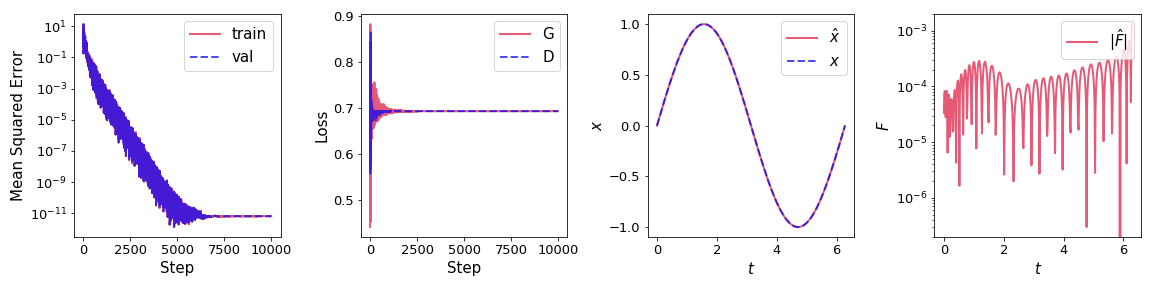}
  \caption[DEQGAN Training: Simple Oscillator]{Visualization of DEQGAN training for the simple harmonic oscillator problem. The left-most figure plots the mean squared error vs. step (iteration) count. To the right of this, we plot the value of the generator (G) and discriminator (D) losses for each step. Right of this we plot the prediction of the generator $\hat{x}$ and the true analytic solution $x$ as functions of time $t$. The right-most figure plots the absolute value of the residual of the predicted solution $\hat{F}$.}
  \label{fig:gan_sho}
\end{figure}

Here we set $LHS=\ddot{x} + x$ and $RHS = 0$. Figure \ref{fig:gan_sho} plots the results of training DEQGAN on this problem. 

% NONLINEAR OSCILLATOR
\paragraph{Nonlinear Oscillator (NLO)}

Further increasing the complexity of the differential equations being considered, consider a less idealized oscillating body subject to additional forces, whose motion $x(t)$ we can described by the nonlinear oscillator differential equation
\begin{equation}
    \ddot{x}(t)+2 \beta \dot{x}(t)+\omega^{2} x(t)+\phi x(t)^{2}+\epsilon x(t)^{3} = 0,
\end{equation}
with $\beta=0.1, \omega=1, \phi=1, \epsilon=0.1$, $x(0) = 0$, $\dot{x}(0) = 0.5$, and $t \in (0, 4\pi)$. This equation does not admit an analytical solution. Instead, we use the high-quality solver provided by SciPy's \url{solve_ivp} \cite{scipy_cite}.

\begin{figure}[h]
  \centering
  \includegraphics[width=\textwidth]{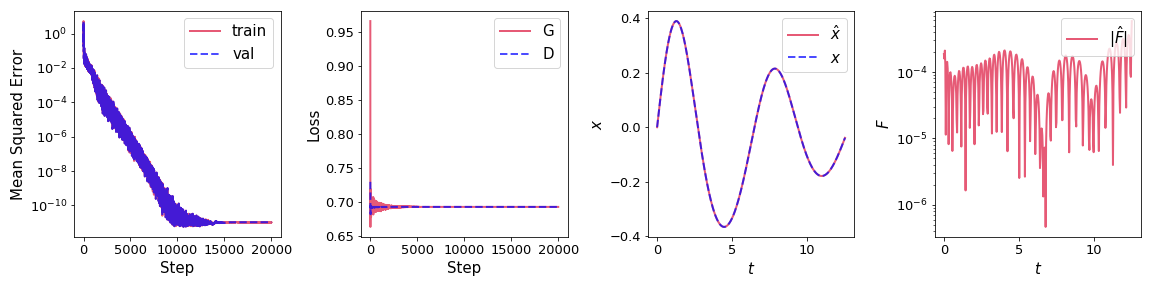}
  \caption[DEQGAN Training: Nonlinear Oscillator]{Visualization of DEQGAN training for the nonlinear oscillator problem. The left-most figure plots the mean squared error vs. step (iteration) count. To the right of this, we plot the value of the generator (G) and discriminator (D) losses for each step. Right of this we plot the prediction of the generator $\hat{x}$ and the ground truth solution $x$ as functions of time $t$. The right-most figure plots the absolute value of the residual of the predicted solution $\hat{F}$.}
  \label{fig:gan_nlo}
\end{figure}

We set $LHS=\ddot{x}+2 \beta \dot{x}+\omega^{2} x+\phi x^{2}+\epsilon x^{3} = 0$ and $RHS=0$. Figure \ref{fig:gan_nlo} plots the results obtained from training DEQGAN on this equation.

% COO / NAS MODEL
\paragraph{Non-Autonomous System (NAS)}

Consider the system of ordinary differential equations given by

\begin{equation}
    \dot{x}(t) = -ty
\end{equation}
\begin{equation}
    \dot{y}(t) = tx
\end{equation}
with $x(0)=1$, $y(0)=0$, and $t \in (0, 2\pi)$. This equation has an exact analytical solution given by

\begin{equation}
    x = \cos\left(\frac{t^2}{2}\right)
\end{equation}
\begin{equation}
    y = \sin\left(\frac{t^2}{2}\right).
\end{equation}

\begin{figure}[h]
  \centering
  \includegraphics[width=\textwidth]{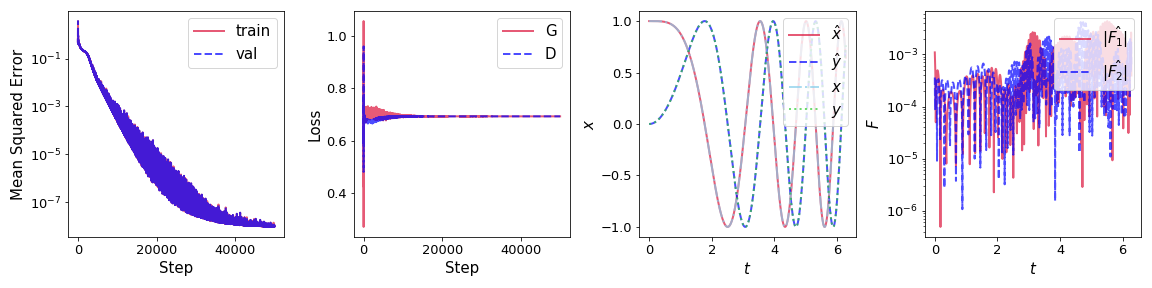}
  \caption[DEQGAN Training: SIR Model]{Visualization of DEQGAN training for the non-autonomous system of equations. The left-most figure plots the mean squared error vs. step (iteration) count. To the right of this, we plot the value of the generator (G) and discriminator (D) losses for each step. Right of this we plot the predictions of the generator $\hat{x}, \hat{y}$ and the true analytic solutions $x$, $y$ as functions of time $t$. The right-most figure plots the absolute value of the residuals of the predicted solution $\hat{F_j}$ for each equation $j$.}
  \label{fig:gan_coo}
\end{figure}

Here we set 
\begin{equation}
    LHS = \left[ \frac{dx}{dt} + ty, \frac{dy}{dt} - xy \right]^T
\end{equation}
and $RHS=\left[0, 0 \right]^T$. Figure \ref{fig:gan_coo} plots the result of training DEQGAN on this problem.

% SIR MODEL
\paragraph{SIR Epidemiological Model (SIR)}

Given the recent outbreak and pandemic of novel coronavirus (COVID-19) \cite{coronavirus}, we consider an epidemiological model of infectious disease spread given by a system of ordinary differential equations. Specifically, consider the Susceptible $S(t)$, Infected $I(t)$, Recovered $R(t)$ model for the spread of an infectious disease over time $t$. The model is defined by a system of three ordinary differential equations

\begin{equation} \label{eq:sir1}
    \frac{dS}{dt} = - \beta \frac{IS}{N}
\end{equation}
\begin{equation} \label{eq:sir2}
    \frac{dI}{dt} = \beta \frac{IS}{N} - \gamma I
\end{equation}
\begin{equation} \label{eq:sir3}
    \frac{dR}{dt} = \gamma I
\end{equation}
where $\beta=3, \gamma=1$ are given constants related to the infectiousness of the disease, $N = S+I+R$ is the (constant) total population, $S(0)=0.99, I(0)=0.01, R(0)=0$, and $t \in (0, 10)$. Again we use SciPy's \url{solve_ivp} solver \cite{scipy_cite} to obtain ground truth solutions.

\begin{figure}[h]
  \centering
  \includegraphics[width=\textwidth]{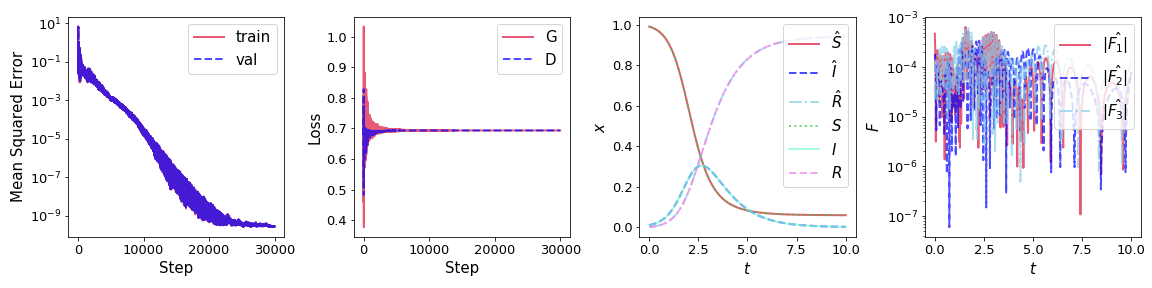}
  \caption[DEQGAN Training: SIR Model]{Visualization of DEQGAN training for the SIR system of equations. The left-most figure plots the mean squared error vs. step (iteration) count. To the right of this, we plot the value of the generator (G) and discriminator (D) losses for each step. Right of this we plot the predictions of the generator $\hat{S}, \hat{I}, \hat{R}$ and the ground truth solutions $S$, $I$, $R$ as functions of time $t$. The right-most figure plots the absolute value of the residuals of the predicted solution $\hat{F_j}$ for each equation $j$.}
  \label{fig:gan_sir}
\end{figure}

We set $LHS$ to be the vector
\begin{equation}
    LHS = \left[ \frac{dS}{dt} + \beta \frac{IS}{N}, \frac{dI}{dt} - \beta \frac{IS}{N} + \gamma I, \frac{dR}{dt} - \gamma I \right]^T
\end{equation}
and $RHS=\left[0, 0, 0 \right]^T$. We present the results of training DEQGAN to solve this system of differential equations in Figure \ref{fig:gan_sir}.

% POISSON EQUATION
\paragraph{Poisson Equation (POS)} 

\begin{figure}[h]
  \centering
  \includegraphics[width=\textwidth]{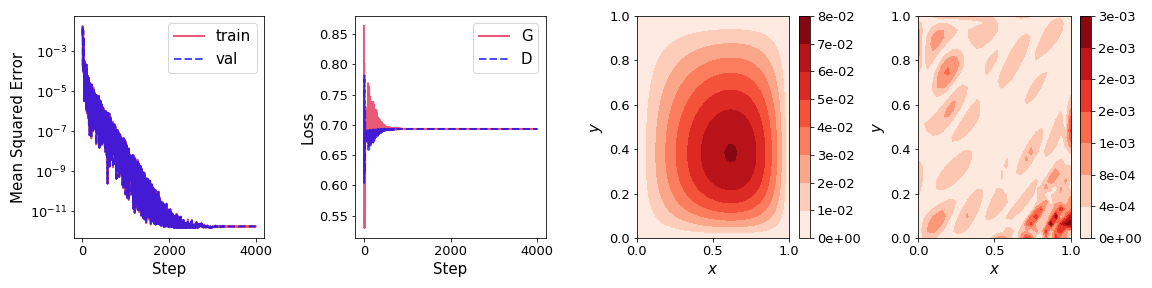}
  \caption[DEQGAN Training: Poisson Equation]{Visualization of DEQGAN training for the Poisson equation. The left-most figure plots the mean squared error vs. step (iteration) count. To the right of this, we plot the value of the generator (G) and discriminator (D) losses for each step. Right of this we plot the prediction of the generator $\hat{u}$ as a function of position $(x,y)$. The right-most figure plots the absolute value of the residual $\hat{F}$, as a function of $(x,y)$.}
  \label{fig:gan_pos}
\end{figure}

Consider the Poisson partial differential equation (PDE) given by \begin{equation}
    \frac{\partial^2 u}{\partial x^2} + \frac{\partial^2 u}{\partial y^2} = 2x(y-1)(y-2x+xy+2)e^{x-y}
\end{equation}
where $(x,y) \in [0, 1] \times [0, 1]$. The equation is subject to Dirichlet boundary conditions on the edges of the unit square \begin{equation} \begin{split}
    u(x, y)\bigg|_{x=0} &= 0 \\
    u(x, y)\bigg|_{x=1} &= 0 \\
    u(x, y)\bigg|_{y=0} &= 0 \\
    u(x, y)\bigg|_{y=1} &= 0. \\
\end{split} \end{equation} The analytical solution is \begin{equation}
    u(x, y) = x(1-x)y(1-y)e^{x-y}.
\end{equation} We use the two-dimensional Dirichlet boundary adjustment formulae provided in \citet{feiyu_neurodiffeq}. To set up the problem for DEQGAN we let \begin{equation}
    LHS = \frac{\partial^2 u}{\partial x^2} + \frac{\partial^2 u}{\partial y^2} - 2x(y-1)(y-2x+xy+2)e^{x-y}
\end{equation}
and $RHS = 0$. We present the results of training DEQGAN on this problem in Figure \ref{fig:gan_pos}.

% Hyperparameters
\subsection*{DEQGAN Hyperparameters}

We performed $1000$ iterations of random search to tune the hyperparameters of DEQGAN for each differential equation. Table \ref{table:hyperparameters} summarizes the final hyperparameter settings used for DEQGAN.

\begin{table}
\caption{Hyperparameter Settings for DEQGAN}
\label{table:hyperparameters}
\begin{center}
\begin{small}
\begin{sc}
\begin{tabular}{lcccccc}
\toprule
Hyperparameter & EXP & SHO & NLO & NAS & SIR & POS  \\
\midrule
Num. Iterations & \num{2e3} & \num{1e4} & \num{2e4} & \num{5e4} & \num{3e4} & \num{4e3} \\
Num. Grid Points & $100$ & $400$ & $400$ & $800$ & $800$ & $32 \times 32$\\
$G$ Units/Layer & $30$ & $40$ & $40$ & $30$ & $40$ & $40$ \\
$G$ Num. Layers & $2$ & $4$ & $4$ & $3$ & $2$ & $4$ \\
$D$ Units/Layer & $20$ & $40$ & $30$ & $50$ & $20$ & $20$ \\
$D$ Num. Layers & $4$ & $2$ & $3$ & $2$ & $3$ & $4$ \\
Activations & $\tanh$ & $\tanh$ & $\tanh$ & $\tanh$ & $\tanh$ & $\tanh$ \\
$G$ Learning Rate & $0.008$ & $0.009$ & $0.006$ & $0.006$ & $0.010$ & $0.008$ \\
$D$ Learning Rate & $0.0005$ & $0.002$ & $0.0007$ & $0.001$ &  $0.002$ & $0.002$ \\
$G$ $\beta_1$ (Adam) & $0.671$ & $0.444$ & $0.102$ & $0.706$ & $0.207$ & $0.410$ \\
$G$ $\beta_2$ (Adam) & $0.143$ & $0.633$ & $0.763$ & $0.861$ & $0.169$ & $0.447$ \\
$D$ $\beta_1$ (Adam) & $0.866$ & $0.271$ & $0.541$ & $0.538$ & $0.193$ & $0.593$ \\
$D$ $\beta2$ (Adam) & $0.165$ & $0.142$ & $0.677$ & $0.615$ & $0.617$ & $0.915$ \\
Exponential LR Decay ($\gamma$) & $0.991$ & $0.998$ & $0.999$ & $0.9998$ & $0.9996$ & $0.996$ \\
\bottomrule
\end{tabular}
\end{sc}
\end{small}
\end{center}
\end{table}

% NON-GAN Tuning
\subsection*{Non-GAN Hyperparameter Tuning}
\label{appendix:non-gan-tuning}

Table \ref{table:experimental_results_non_gan_tuning} presents results after tuning the hyperparameters of the alternative unsupervised neural network method with $L_1$, $L_2$, and Huber loss functions. We ran $1000$ random search iterations for each differential equation.

\begin{table}[h]
  \caption{Experimental Results With Non-GAN Hyperparameter Tuning}
  \label{table:experimental_results_non_gan_tuning}
  \centering
  \begin{tabular}{lccccccccc}
    \toprule
    & \multicolumn{5}{c}{Mean Squared Error} \\ 
    \cmidrule(lr){2-6}
    Key & $L_1$ & $L_2$ & Huber & DEQGAN & Traditional  \\ 
    \midrule
    EXP & \num{1e-4} & \num{3e-8} & \num{1e-8}  & \num{3e-16} & \num{2e-14} (RK4) \\
    SHO & \num{2e-5} & \num{3e-9} & \num{1e-9} & \num{1e-12} & \num{1e-11} (RK4) \\
    NLO & \num{2e-5} & \num{5e-10} & \num{6e-10} & \num{2e-12} & \num{4e-11} (RK4) \\
    NAS & \num{5e-1} & \num{2e-4} & \num{7e-6} & \num{8e-9} & \num{2e-9} (RK4) \\
    SIR & \num{2e-5} & \num{6e-10} & \num{3e-10} & \num{2e-10}  & \num{5e-13} (RK4) \\
    POS  & \num{1e-5} & \num{3e-10} & \num{2e-10} & \num{8e-13} & \num{3e-10} (FD) \\
    \bottomrule
  \end{tabular}
\end{table}

\end{document}